# Millisecond-Response Tracking and Gazing System for UAVs: A Domestic Solution Based on "Phytium + Cambricon"

Zhu Yuchen[1], Yin Longxiang[2], Zhao Kai[3]

*(1.Beijing Information Science and Technology University, Beijing 102206, China; 2. Institute of Computing Technology, Chinese Academy of Sciences, Beijing 100190, China; 3.Beijing Information Science and Technology University, Beijing 102206, China)*

**Abstract:** In the frontier research and application of current video surveillance technology, traditional camera systems exhibit significant limitations of response delay exceeding 200 ms in dynamic scenarios due to the insufficient deep feature extraction capability of automatic recognition algorithms and the efficiency bottleneck of computing architectures, failing to meet the real-time requirements in complex scenes. To address this issue, this study proposes a heterogeneous computing architecture based on Phytium processors and Cambricon accelerator cards, constructing a UAV tracking and gazing system with millisecond-level response capability. At the hardware level, the system adopts a collaborative computing architecture of Phytium FT-2000/4 processors and MLU220 accelerator cards, enhancing computing power through multi-card parallelism. At the software level, it innovatively integrates a lightweight YOLOv5s detection network with a DeepSORT cascaded tracking algorithm, forming a closed-loop control chain of "detection-tracking-feedback". Experimental results demonstrate that the system achieves a stable single-frame comprehensive processing delay of 50-100 ms in 1920×1080 resolution video stream processing, with a multi-scale target recognition accuracy of over 98.5%, featuring both low latency and high precision. This study provides an innovative solution for UAV monitoring and the application of domestic chips.

# 1 Introduction

In recent years, with the continuous innovation of science and technology and the gradual enhancement of public safety awareness, video surveillance systems have become important tools for maintaining public security and promoting social management.

In the field of visual tracking and gazing technology, fast response performance is a core element ensuring system effectiveness. Its impact on target tracking effects manifests in multiple aspects. As shown in Figure 1, system response delays can trigger a series of performance degradation issues, which seriously affect the performance and accuracy of tracking and gazing. [1]These mainly include the following:

a) Target loss: In cases of slow response, if the target moves or changes significantly while the system is processing the previous video frame, the tracking algorithm may fail to redetect and recognize the target in subsequent frames, leading to target loss and thus affecting tracking and gazing results.

b) Motion prediction issues: Slow system response may result in incorrect estimation of the target's movement speed and direction, as some tracking algorithms may predict the target's next movement direction and position based on incomplete information, which will affect the algorithm's motion prediction of the target.

c) Impact on real-time performance: In real-time tracking systems, slow video frame processing speed leads to slow overall system response, thereby affecting the system's real-time response capability.

d) Data association issues: Slow response may make it difficult to correctly associate targets in the current frame with target information from previous frames during tracking.

e) Tracking interruption: Slow response causes the system to be unable to process the large number of consecutive video frames, which may result in the tracking algorithm failing to maintain target continuity between consecutive frames, leading to tracking interruption.

f) Misalignment: Slow response may cause the tracking algorithm to misalign different targets, especially in multi-target tracking scenarios where the appearance or motion characteristics of different targets may become blurred due to slow response.

g) Tracking drift: Due to slow response, the tracking algorithm may fail to update the target's state in a timely manner, causing the tracking frame to gradually deviate from the target's actual position.

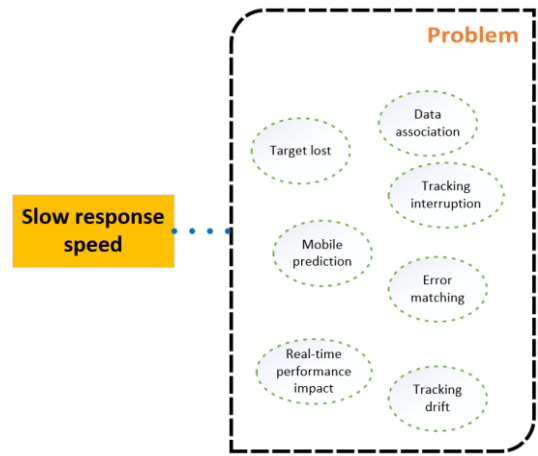

Figure 1 The impact of slow response speed on tracking gaze

In the field of tracking and gazing, the slow response speed of the entire system is usually attributed to two main factors: hardware performance limitations and algorithm complexity issues. In some cases, the execution of tracking and gazing tasks depends on the computing power of the hardware platform. If the hardware computing power is insufficient to support the required processing speed, it may be impossible to process all incoming video frames in real-time, resulting in slow response times. Additionally, the efficiency of the tracking algorithm itself is a key factor affecting response time. If the algorithm is not optimized enough or has bottlenecks in computational complexity, even high-performance hardware may fail to achieve efficient processing. In tracking systems requiring real-time feedback, algorithms must complete processing each frame in an extremely short time and promptly feed back control signals to the equipment. If the algorithm's time complexity is too high, it may fail to meet real-time processing requirements, leading to slow response. Therefore, to solve this problem, both improving hardware computing power and optimizing algorithms must be considered to ensure the tracking system operates stably under various conditions while maintaining efficient and accurate tracking performance.

Maintaining the continuity and integrity of video frame output is a key prerequisite for ensuring target tracking and gazing accuracy, which affects performance in specific application scenarios. Fast-response tracking and gazing systems can be applied to assist maritime law enforcement vessels, helping law enforcement fleets effectively and accurately identify various specific targets and continuously track them to combat illegal sand mining, illegal sea use, and other behaviors damaging marine resources, as well as criminal activities such as smuggling and human trafficking. Beyond such scenarios, multiple other situations also require target tracking and gazing technology with high computing power and efficient algorithms to avoid potential problems caused by delayed responses:

a) Industrial automation and intelligent robots: Industrial machinery needs to accurately identify goods and assembly parts, and robots on production lines need to recognize and track products for sorting, assembly, etc.

b) Sports event analysis: In sports events such as basketball, football, and table tennis, high-computing-power tracking and gazing technology can track athletes and balls for motion analysis, foul judgment, and replay.

c) Military field: In complex battlefield environments, high-computing-power tracking technology is used to track enemy targets for intelligence collection and missile guidance.

d) Traffic management: In traffic monitoring systems, tracking vehicle flow for traffic analysis and violation detection.

Faced with massive data scales and demands for fast responses, slow response issues often arise, which must be addressed with more intelligent and powerful computing capabilities. The tracking and gazing system proposed in this study integrates target detection and tracking algorithms on a PTZ camera and relies on the powerful computing power of Phytium processors and Cambricon AI acceleration chips to achieve accurate target recognition and real-time tracking, effectively solving response delay issues caused by insufficient computing power or algorithms. The tracking and gazing system integrates technological innovation, practical application needs, theoretical deepening, and socio-economic development requirements. With significant advancements in computer vision and machine learning, especially breakthroughs in deep learning, strong technical support is provided for accurate and robust target tracking. In key fields such as security monitoring, autonomous driving, intelligent traffic management, and industrial automation, demands for technologies capable of accurately identifying, locating, and tracking targets continue to grow. Additionally, improved hardware performance, such as the popularization of high-resolution cameras and high-speed processors, enables real-time processing of large amounts of video data.

Through precise target detection and tracking algorithms, the system achieves real-time monitoring of specific targets in dynamic scenarios, greatly improving monitoring efficiency and response speed. The system can be applied in numerous fields. In security monitoring, it can effectively prevent and promptly respond to various security threats, providing strong technical support for public security and border monitoring. In healthcare, through surgical navigation and telemedicine, it can provide doctors with more accurate surgical assistance, improving success rates. In industrial production, its target recognition and tracking capabilities in automated production lines promote efficiency and reduce costs. Socio-economically, it not only drives the development of related technologies but also provides impetus for the digital transformation of the economy, boosting productivity and innovating business models. [2]

## 2 Algorithm Optimization and Accelerator Card Parallelism: Software and Hardware Solutions

During target tracking and gazing tasks, system response delay is a key bottleneck restricting performance improvement, with non-negligible impacts. This is mainly due to the complexity of algorithm architectures, insufficient optimization, and reliance on hardware platforms inadequate for required processing speeds. Traditional target tracking algorithms have design limitations and lack targeted optimization; moreover, their hardware platforms typically rely solely on CPUs or rarely utilize acceleration units for computing module acceleration, making them struggle to achieve ideal results in processing high-complexity target tracking tasks. [3] The system proposed in this study improves existing technical solutions from both hardware architecture reconstruction and software algorithm optimization dimensions, enhancing system performance and tracking accuracy, and effectively solving slow response issues in tracking and gazing. Figure 2 shows the software and hardware solutions designed by the system to address this problem.

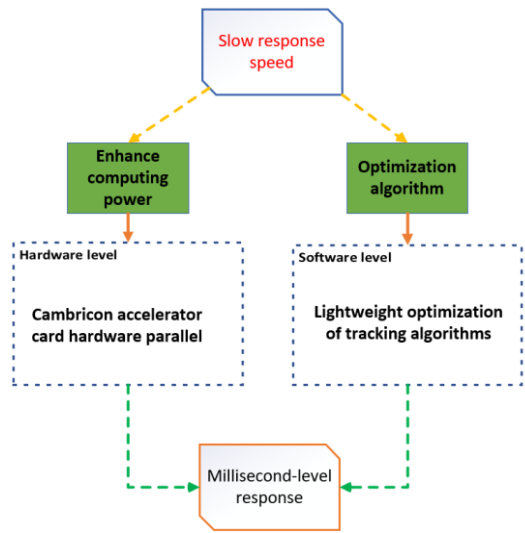

Figure 2 Software and hardware solutions for slow response speed issues

As seen in Figure 2, the system primarily addresses slow response from two aspects: enhancing computing power and optimizing algorithms. On one hand, it improves computing power at the hardware level by using domestic Cambricon high-performance accelerator cards for hardware acceleration of algorithm computing modules, increasing computing power, and achieving linear growth in computing power through multi-card collaboration. [4] On the other hand, it performs more lightweight optimization of algorithms to reduce complexity and achieve faster processing speeds. The system combines hardware acceleration and algorithm optimization to bidirectionally improve computational efficiency in target tracking, ultimately reducing response speed to the millisecond level. Subsequent chapters will detail the working mechanism and key technical principles of the system's hardware architecture. [5]

## 3 Collaborative Working Mechanism of Phytium Processors and Cambricon Accelerator Cards

### 3.1 Cambricon Edge AI Accelerator Card MLU220

The Cambricon MLU220 (Machine Learning Unit 220), a dedicated AI acceleration chip for edge computing scenarios, belongs to edge AI accelerator cards. With its small size and high intelligence, it shows broad application prospects in fields such as intelligent robots, smart factories, smart finance, and smart retail. The Cambricon Intelligent Edge Computing Module MLU 220 (Siyuan 220) chip, based on its advanced MLUv02 architecture, integrates 8TOPS theoretical peak performance with a power consumption of only 8.25W. Its high energy efficiency makes it ideal for terminal and edge computing devices, providing powerful AI processing capabilities. Table 1 lists the specifications of MLU 220-M.2 in terms of AI performance, memory, encoding/decoding capabilities, image decoding, and power consumption. [6]

Table 1　MLU220-M.2 specification parameters

| Parameter | Specification |
| --- | --- |
| AI Performance | 8TOPS（INT8） |
| Memory | LPDDR4x 64 bits |
| Encoding/Decoding Capability | H.264，HEVC (H.265)，VP8，VP9； |
| Image Decoding | JPEG，maximum resolution 8192×8192 |
| Power Consumption | 8.25W |

The MLU220 chip has three core advantages:

a) Data privacy protection:

The Cambricon MLU220 edge AI accelerator card adopts localized data processing, achieving integrated computing-storage through on-chip storage (64-bit LPDDR4x), avoiding cloud transmission of sensitive data (such as video streams), and reducing data leakage risks by over 90%.

b) Low processing latency:

The MLU220 chip's high-performance computing capabilities enable fast responses, realizing real-time or near-real-time data processing and improving overall system response speed. Low latency is crucial for application scenarios requiring rapid decision-making, such as autonomous driving and industrial automation.

c) High bandwidth utilization:

The MLU220 chip efficiently utilizes available bandwidth, ensuring maximum utilization of data transmission and reducing performance bottlenecks caused by bandwidth limitations. High bandwidth utilization means the chip can maintain high throughput when processing large amounts of data, improving data processing efficiency. [7]

### 3.2 System Hardware Architecture

As a domestic CPU, Phytium processors have powerful computing capabilities, while Cambricon chips are dedicated to AI computing. Their combination enables efficient model deployment, computing, and inference acceleration. The system's hardware adopts Phytium processors integrated with Cambricon MLU hardware accelerator cards for collaborative computing, real-time processing of video stream data from the UAV PTZ camera via video channels, performing deep learning model inference, and guiding the PTZ camera through serial communication to support fast target detection and tracking. [8]

Phytium processors serve as the system's main control unit, responsible for global scheduling of computing tasks and resource coordination, while Cambricon MLU accelerator cards act as the system's computing engine, executing complex algorithms and data processing tasks. Through data transmission channels, Phytium processors offload complex algorithms and data computations on video data to MLU accelerator cards. This ensures the system can process high-resolution video streams and real-time data, meeting millisecond-level response requirements and improving target recognition accuracy and efficiency. [9] Figure 3 shows the system's hardware architecture.

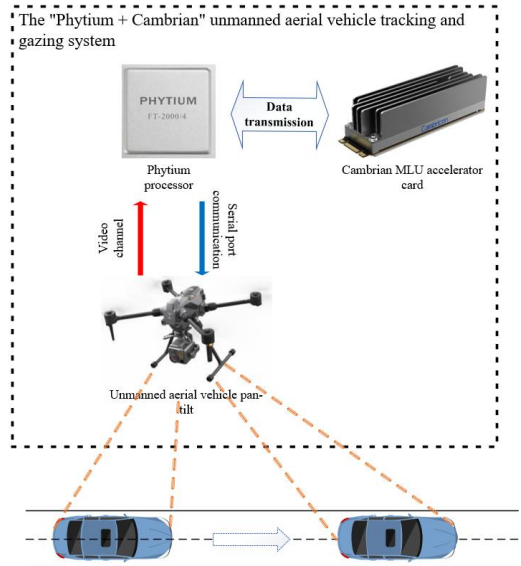

Figure 3 Hardware Architecture of the UAV Tracking and Gazing System

In image capture, Phytium processors can quickly process data streams from cameras, performing initial data processing and encoding to prepare for subsequent image processing. Cambricon MLU, through its dedicated AI processor cores, can efficiently execute image filtering, edge detection, feature extraction, and other operations. The UAV PTZ camera is responsible for perception, capturing motion video images of objects, and accepting processor control commands to rotate, achieving tracking and gazing functions.

In the design of the tracking and gazing system, Phytium CPUs and Cambricon AI accelerator cards MLU220 are closely connected via high-speed PCIE interfaces, ensuring efficient data transmission and processing. The system also includes necessary storage and network components to support large-scale data processing. Specifically, this hardware configuration allows the system to fully utilize Phytium CPU's computing power and Cambricon MLU220 accelerator card's AI inference performance when executing complex tracking and gazing tasks. The high-bandwidth characteristic of PCIE interfaces enables fast data flow between processors and accelerator cards, reducing potential bottlenecks and improving overall processing speed. Through this comprehensive and stable hardware architecture, the tracking and gazing system achieves more accurate and reliable target recognition and tracking analysis.

**3.3 System Workflow**

This system primarily studies the recognition and tracking of single targets, i.e., identifying a specific set target, which can be extended to multi-target tracking and gazing. The system's workflow is shown in Figure 4. The specific steps are:

1.The UAV PTZ camera captures video images of the target.

2.The Phytium CPU preprocesses the images, including improving image quality, enhancing features, and converting video image formats to resolutions suitable for the system to facilitate subsequent processing. [10]

3.The Cambricon MLU performs image inference, running target detection algorithms to identify targets and display detection boxes and target confidence.

4.After the target moves, the Phytium CPU runs target tracking algorithms, combines the identified target, controls the camera to rotate and follow the target's movement, and continuously tracks and gazes at the target.

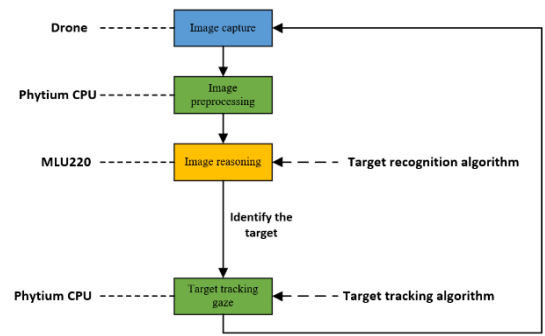

Figure 4 Schematic diagram of system operation

**4 Small and Medium Target Recognition Algorithm Based on YOLOv5s**

**4.1 YOLOv5 Algorithm**

The system uses the YOLOv5 (You Only Look Once version 5) target detection algorithm. As a target detection algorithm under the deep learning framework, YOLOv5 is characterized by an efficient balance between detection speed and accuracy. Its core idea is to treat target detection as a single regression problem, enabling real-time detection and balancing speed and accuracy by using convolutional neural networks to directly predict object bounding boxes and category labels. It has many advantages over other target detection algorithms, specifically:

a) Speed optimization: The system needs to promptly process video streams transmitted from PTZ cameras to quickly take next actions. YOLOv5 achieves faster detection speeds while maintaining high accuracy, making it very suitable for real-time video stream processing and applications requiring fast responses.

b) Model fine-tuning and optimization: YOLOv5 allows flexible control of network depth and width through depth_multiple and width_multiple parameters, adapting to different application scenarios and resource constraints.

c) Easy training and deployment: YOLOv5 provides pre-trained models and simple training interfaces, supporting deployment on multiple platforms and devices, including CPUs, GPUs, and mobile edge devices.

d) Scalability: YOLOv5 supports models of different sizes, allowing users to select appropriate model versions based on needs, balancing detection speed and accuracy.

e) Multi-scale detection: YOLOv5 can accurately detect targets of different scales, shapes, and poses, with good adaptability.

Given the above advantages of YOLOv5 selecting this algorithm as the system's target detection solution is appropriate. [11] Considering the system needs to integrate with edge PTZ camera devices and meet real-time response requirements, the small and medium-sized model YOLOv5s, which is suitable for edge devices and has the fastest detection speed, is chosen.

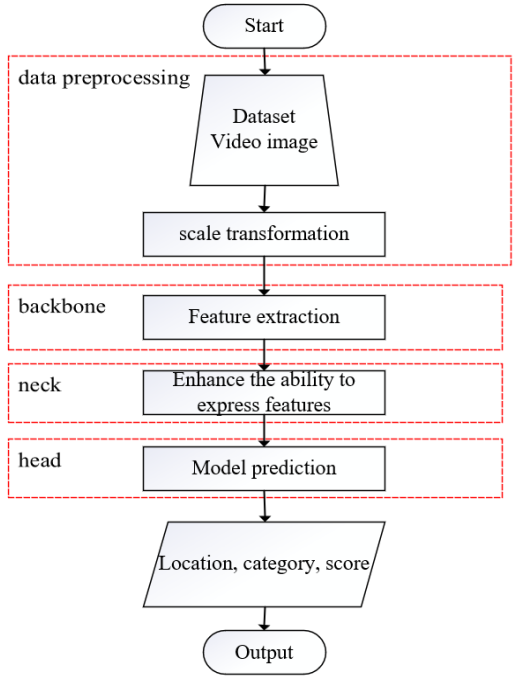

Figure 5 Schematic diagram of YOLOv5 model processing flow

As shown in Figure 5, in the YOLOv5 model processing flow, input images first undergo a series of preprocessing steps, including pixel value normalization, image size adjustment, and anchor box generation based on predefined parameters, ensuring input data consistency to enhance model accuracy and robustness. Preprocessed images are then sent to the backbone network for feature extraction, which uses an improved CSPDarknet53 network structure to ensure high detection accuracy, capture rich feature representations, and minimize computational load and memory usage. Next, feature information from the backbone layer is further processed through the neck network. The neck network uses a PANet structure for multi-scale feature fusion, achieving effective integration of features at different levels through multiple upsampling, feature concatenation, and dot product operations, which helps the model capture targets at different scales. After enhancing feature expression capabilities through the neck layer, features extracted by the backbone are passed to the head layer for final model prediction, i.e., detecting target positions and categories using feature maps extracted by the backbone network. The final output is the model's prediction results, including each target's category and corresponding bounding box coordinates. Figure 6 shows the original YOLOv5 network structure.

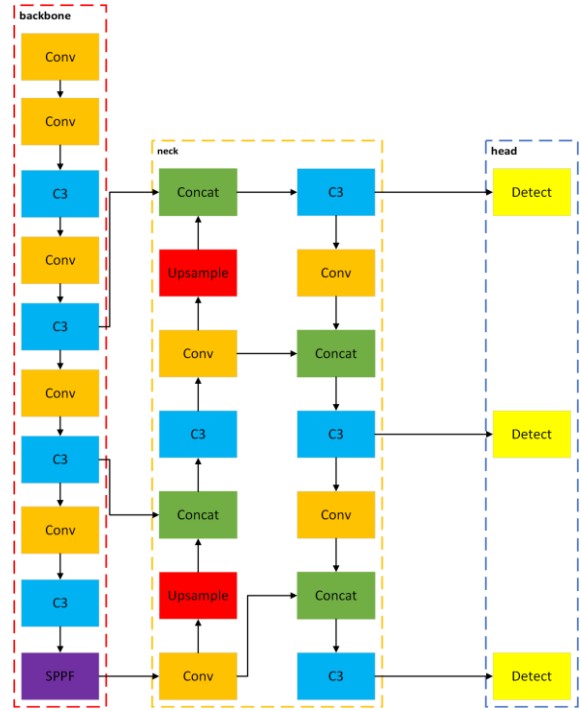

Figure 6 The network structure of YOLOv5

### 4.2 Improved YOLOv5

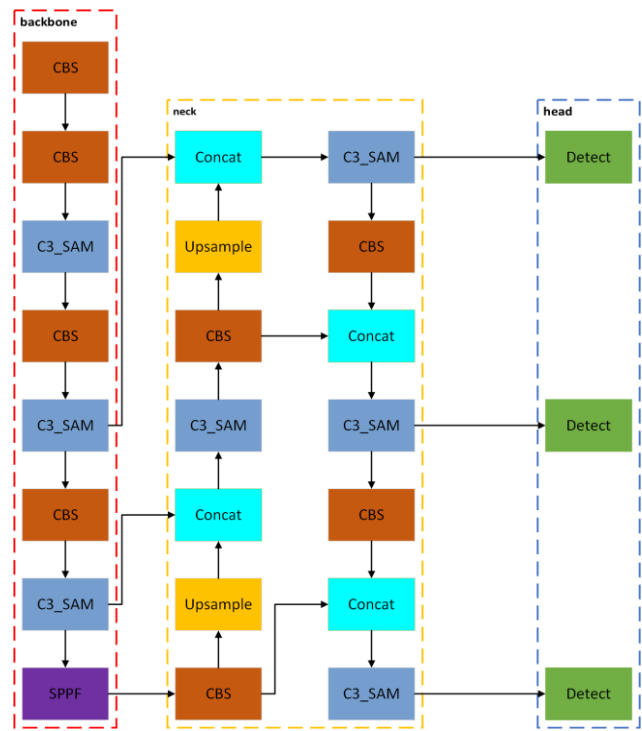

Figure 7 Improved network architecture

YOLOv5 is mainly suitable for identifying general-sized small and medium objects. In this system, since specific targets are identified, directly using the traditional YOLOv5 network model for target detection may lead to loss of small target details and unnecessary consumption of computing resources when detecting large targets. Therefore, optimization and improvement of the YOLOv5s model are carried out, focusing on network depth adjustment and structural module reconstruction. By designing a backbone network and neck network specifically for certain small targets, subtle features of targets can be extracted more efficiently, thereby improving network detection accuracy and achieving faster detection speeds.

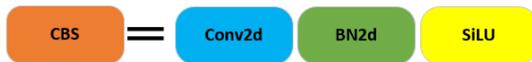

Figure 8 The structure of CBS module

Although deep networks can capture deeper semantic features and show advantages in processing large-scale datasets, they may lose target spatial information, making it difficult to capture specific details during target localization. Therefore, to solve the problem of algorithm accuracy degradation caused by insufficient target detail information, the original backbone network is optimized. The improved YOLOv5 structure is shown in Figure 7. To address potential loss of small target information, a CBS module is first introduced, whose structure is detailed in Figure 8. Specifically, the Conv modules shown in Figure 6 in the original YOLOv5 network structure are replaced with CBS modules. This improvement aims to enhance the network's ability to capture small target features. By introducing CBS modules, the network can learn more complex feature representations, thereby enhancing feature extraction capabilities. The CBS module mainly consists of three parts: a convolutional layer (Conv), a batch normalization layer (BN), and an activation function (SiLU). The convolutional layer here is the same as the Conv in the original YOLOv5 network structure, responsible for convolving input feature maps to extract image features. The batch normalization layer normalizes the output of the convolutional layer, making the data conform to a certain distribution, thereby accelerating model convergence and improving model stability and generalization ability. The SiLU activation function introduces non-linear factors, enabling the model to learn more complex mapping relationships. The SiLU activation function is shown in formula (1), which is a variant of the swish activation function with smooth and non-monotonic characteristics, helping to alleviate the gradient vanishing problem.

$$silu(x) = x * sigmoid(x) = \frac{x}{1+e^{-x}} \quad (1)$$

The CBS module realizes feature extraction and non-linear transformation of input feature maps by combining convolutional layers, batch normalization layers, and activation functions. In the YOLOv5 algorithm, a backbone network with strong feature extraction capabilities can be constructed by stacking multiple CBS modules, as well as extracting deeper features of targets in subsequent neck networks.

In addition, a Spatial Attention Mechanism (SAM) is introduced into the C3 module of the original YOLOv5, making the network pay more attention to pixel regions that are decisive for classification, as shown in Figure 9. This mechanism assigns different weights to each spatial position in the feature map, emphasizing the position information of the target object, thereby enhancing attention to small target regions. After introducing SAM, the C3 module can converge faster because it can more effectively focus on features useful for target detection, reducing the search space during model training. SAM enhances the position information of target objects by assigning different weights to different spatial positions, thereby improving network detection accuracy.

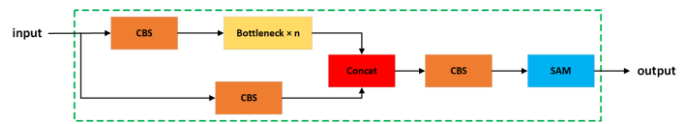

Figure 9 The structure of C3 module

To improve model training efficiency and detection and localization accuracy, the SIoU loss function (Shape-Aware IoU) suitable for high-precision bounding box localization scenarios is introduced. SIoU is an advanced bounding box regression loss function that optimizes bounding box localization accuracy by combining angle cost, distance cost, shape cost, and IoU cost, accelerating model convergence and improving model detection and localization accuracy. The SIoU loss function is expressed as formula (2), where cd represents distance cost, cs represents shape cost, and both eps and α are adjustable parameters. [12]

$$SIoU = IoU - (0.5 \times (c_d + c_s) + eps)^\alpha \quad (2)$$

Finally, the HardTanh activation function is used instead of the LeakyReLU activation function. LeakyReLU (Leaky Rectified Linear Unit) is an improved linear activation function provided by PyTorch, a variant of ReLU, whose function formula can be expressed as
{{f(x)=}\left\{{\begin{eqnarray}{{&x, &&ifx≥0}}\\{{&ax, &&ifx<0 &&}}\\\end{eqnarray}}\right.}. Compared with ReLU, LeakyReLU is characterized by not directly setting negative inputs to zero but allowing them to pass through with a non-zero slope. This parameter is usually a small positive number, enabling negative input values to enter the network with a small amplitude instead of being completely blocked. This design helps solve the "dying neuron" problem that may occur during ReLU training, where ReLU output remains zero when input is continuously negative, leading to gradient vanishing and neurons no longer updating. LeakyReLU keeps neurons active even for negative inputs by introducing this small slope, facilitating network learning and reducing the possibility of gradient vanishing.

The formula for the HardSwish activation function is formula (3):

$$HardSwish(x) = x \cdot HardSigmoid(x)$$

$$= x \cdot \begin{cases} 1, & \text{if } x \geq 3 \\ \frac{x}{6} + \frac{1}{2}, & \text{if } -3 < x < 3 \\ 0, & \text{if } x \leq -3 \end{cases} \quad (3)$$

After derivation with respect to x, this function becomes a piecewise function. The advantage of piecewise functions is that they can reduce the number of memory accesses, thereby significantly reducing latency. Reducing frequent memory access helps improve processing speed and response time.

Compared with LeakyReLU, it shows multiple advantages: it enhances the model's ability to capture complex data patterns through enhanced non-linear characteristics, and simplifies calculations to reduce model computing costs and improve training and inference efficiency. As a continuously differentiable function, Hardswish is easier to integrate into deep learning frameworks, facilitating cross-platform deployment and optimization of the model. Additionally, in this experiment, considering that the Cambricon MLU220 serves as an edge acceleration device, Hardswish can provide better performance on specific datasets and model structures, making it more adaptable to specific tasks or application scenarios.

The above optimizations aim to improve the performance of the YOLOv5 model in specific target detection tasks, especially in small target detection, by optimizing network structure and parameter settings to improve detection accuracy and efficiency.

### 4.3 Cambricon Model Acceleration

Considering the inference efficiency issue of YOLOv5 on CPUs, the system adopts the Cambricon MLU220 as a dedicated hardware accelerator to achieve faster inference speeds, which requires porting YOLOv5 to MLU for accelerated inference. [13] As shown in Figure 10, the process of porting YOLOv5 to Cambricon MLU is divided into 5 steps, each of which will be introduced.

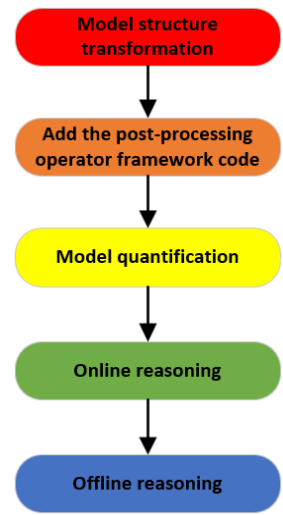

Figure 10 Flowchart of YOLOv5 Model Cambrian Platform Migration

**4.3.1 Model Structure Conversion**

To enhance post-processing performance, the original model architecture of YOLOv5 needs to be adjusted, replacing the original post-processing part in YOLOv5 with BANGC operators provided by Cambricon-PyTorch. This involves modifying the native PyTorch network, converting the computational tasks of the post-processing stage from PyTorch implementation to BANGC operator implementation. Specifically, the yololayer in the model is replaced with the BANGC operator yolov5_detection_output, i.e., the input data originally passed to the three yololayer layers is instead passed to the BANGC operator yolov5_detection_output for processing.

### 4.3.2 Adding Post-Processing Operator Framework Code

In the previous step, the Yolov5Detection operator written in BANGC was used to replace the original post-processing logic of YOLOv5. To ensure correct invocation of this operator, the YOLOv5Detection operator needs to be integrated into the framework. This is divided into two steps: first, integrating the operator into CNPlugin (a collection of BANGC operators launched by Cambricon, mainly based on the function interface provided by CNML to facilitate users to combine MLU operators written in BANG C with CNML and the framework for running and calling), and then integrating the CNPlugin operator into Cambricon-Pytorch. The specific steps for integrating the operator into Cambricon-Pytorch are as follows:

a) Declare the operator in the yaml file
b) Add the CPU implementation of the operator in the OpMethods base class
c) Add a wrapper

The wrapper is an encapsulation layer for underlying operators (kernels), providing a corresponding interface for each operator, with one wrapper per operator. Inference or training operators are preferentially distributed to the wrapper. Add the wrapper implementation according to the template-generated wrapper header file cnml_kernel.h.

d) Add a kernel

The wrapper implements operator functions by calling the kernel. The specific implementation of the operator is mainly completed by calling the CNML library. Figure 11 is a brief logic flowchart of the CNML library. The kernel implementation is completed by calling the CNML library interface according to the programming logic in Figure 11.

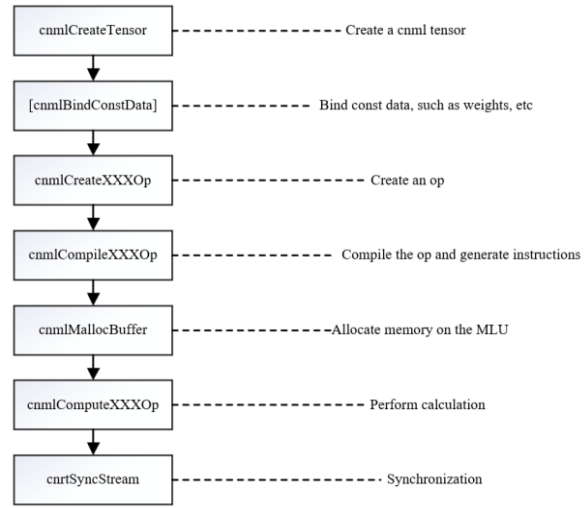

Figure 11 Logic flowchart of CNML library

e) Recompile Cambricon-Pytorch and enter the python environment to confirm successful integration.

### 4.3.3 Model Quantization

Quantization is a deep learning model optimization technology that converts FLOAT32 models to INT8/INT16 models, which can significantly reduce model storage space and bandwidth usage, lower computational complexity, and accelerate inference while ensuring computational accuracy within the target error range. The model acceleration inference of this system uses INT8 quantization to quantize existing model files and generate new quantized models. INT8 models refer to saving values as signed 8-bit integer data and providing the exponent position and scaling factor of int8 fixed-point numbers. The device only supports inputting quantized models during online inference and offline model generation.

### 4.3.4 Online Inference

Perform inference tests on the quantized model generated in the previous step, inferring on images or video frames, and displaying target detection boxes and object confidence. The inference process mainly has the following two modes:

Fusion mode: Multiple fused layers run as a single operation (single Kernel) on the MLU.

According to whether layers in the network can be fused, the network is split into several sub-network segments. Data copying between MLU and CPU only occurs between each sub-network.

Layer-by-layer mode: Each layer operation runs as a single operation (single Kernel) on the MLU, and users can export each layer's results to the CPU for easy debugging.

**4.3.5 Offline Inference**

Offline inference refers to batch processing pre-collected data without real-time requirements. Its process is: first generate an offline model, pre-collect data, perform data preprocessing, and finally conduct offline inference. Generating an offline model is similar to online inference code, and this system mainly uses online inference with real-time performance.

**4.3.6 Model Acceleration**

After YOLOv5 is successfully ported to MLU220, the YOLOv5 target detection algorithm can run on MLU. The MLU220 will accelerate the inference of YOLOv5, relying on its high-performance computing capabilities to achieve real-time data processing and computation, enabling the system to respond quickly and achieve efficient real-time target detection. This is specifically reflected in the following aspects: The MLU acceleration unit converts the model's floating-point weights to low-precision integer format INT8, reducing computational complexity and memory usage; MLU supports highly parallel data processing, enabling simultaneous execution of multiple operations to accelerate model computation; Key operators used in YOLOv5 (such as convolution and pooling) are specially optimized to adapt to MLU's hardware characteristics; Ensure interface compatibility between the MLU platform and mainstream deep learning frameworks (such as PyTorch); Support multi-card parallel processing to improve model processing capabilities and meet large-scale data processing needs.

# 5 Tracking and Gazing Based on DeepSORT Algorithm

**5.1 SORT Algorithm**

The SORT (Simple Online and Realtime Tracking) algorithm, as a concise and efficient real-time online target tracking algorithm, can achieve efficient target tracking tasks in resource-constrained environments. The key to the SORT algorithm is that it combines target detection, Kalman filtering, and the Hungarian algorithm to achieve target tracking. It first provides initial target bounding boxes based on target detection algorithms. Common target detection algorithms such as the representative one-stage algorithm YOLO and two-stage algorithm Faster R-CNN can detect targets in video frames and provide target candidate boxes. Since the accuracy of target detection directly affects tracking results, through the comparison between YOLOv5 and Faster R-CNN in previous chapters and combined with actual scenarios, YOLOv5s is selected as the target detection algorithm. After outputting the target detection results, the SORT algorithm uses the Kalman filtering algorithm to estimate the current state of the target (current position, movement direction, and speed). [14]The Kalman filtering algorithm uses a recursive filtering technique, which can accurately predict the possible position of the target in the next frame. Then, SORT uses the Hungarian algorithm to solve the data association problem, matching the target in the current frame with the target in the previous frame to establish the association relationship of the target. This method matches the target in the current frame with the tracked target in the previous frame by minimizing the association cost to determine the consistency of the target. [15]

**5.2 Kalman Filter Algorithm Prediction**

The Kalman filtering algorithm realizes more accurate tracking target estimation based on sensor measurements and Kalman filter predictions, and can predict the next development trend of the system (here referring to the state of the detected target) in any dynamic system containing uncertain information. A classic example of Kalman filtering is predicting the coordinates and speed of an object's position from a sequence of video frames containing interference noise.

The Kalman filter assumes that variables are random and follow a Gaussian distribution, with each variable having a mean μ (representing the most likely state) and a variance σ² (representing uncertainty). In target tracking, the target's position information is represented by the vector , where each component represents: the center coordinates (u, v) of the bounding box, area s, aspect ratio r, and their respective speed changes. The Kalman filter adopts a constant velocity model and a linear observer model. When tracking a target, it is necessary to predict the state of the target at time t+1 as shown in formulas (4) and (5).

$$x_{t+1} = Fx_t \quad (4)$$
$$P_{t+1} = FP_tF^T + Q \quad (5)$$

Among them, F represents the prediction matrix, which in the constant velocity model is expressed as formula (6):

$$F = \begin{bmatrix} 1 & 0 & 0 & dt & 0 & 0 & 0 \\ 0 & 1 & 0 & 0 & dt & 0 & 0 \\ 0 & 0 & 1 & 0 & 0 & dt & 0 \\ 0 & 0 & 0 & 1 & 0 & 0 & dt \\ 0 & 0 & 0 & 0 & 1 & 0 & 0 \\ 0 & 0 & 0 & 0 & 0 & 1 & 0 \\ 0 & 0 & 0 & 0 & 0 & 0 & 1 \end{bmatrix} \quad (6)$$

In addition, P is the covariance matrix as an empirical parameter, and Q is the system noise.

Taking the system's position prediction of a car as an example, as shown in Figure 12, suppose a car equipped with a ranging radar detects an obstacle directly ahead. At time t1, the radar measures the distance from the car to the obstacle ahead as 25m. Assuming the car's speed is 5m/s, after calculation, the distance from the car to the obstacle at time t2 one second later should be 20m. However, in reality, the radar measures the distance as 22m. Thus, at time t2, there are two distance data: 22m measured by the sensor and 20m calculated. Assuming ideal conditions, the accuracy of the radar sensor changes with distance. Therefore, neither can be used alone; instead, a weighted average of the ideal state and the sensor-measured data should be taken. Assuming the sensor measurement is more accurate with 95% confidence, and the theoretical calculation result is relatively less accurate with 90% confidence, the final distance calculation result is as shown in formula (7):

$$s = (1 - \frac{0.95}{0.95+0.9}) \times 20 + \frac{0.95}{0.95+0.9} \times 22 = 21.03m \quad (7)$$

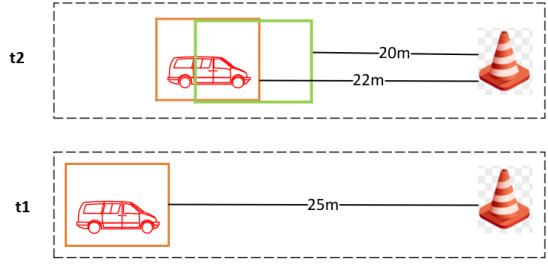

Figure 12 Kalman filtering algorithm for predicting a car

The idea of the Kalman filtering algorithm is to fuse two relatively unreliable methods to obtain a relatively accurate predicted value. It uses the previous moment to predict the next moment and corrects the predicted value with the actual observation at the next moment to obtain the optimal state estimation. In summary, it combines the predicted value calculated by the linear mathematical model with the true value measured by the sensor to obtain a more accurate measured value. The application of the Kalman filtering algorithm in target tracking is to predict the position of the detected target in the next frame based on the position of the detection box.

**5.3 Hungarian Algorithm for Data Association**

The Hungarian algorithm was originally designed to solve the maximum matching problem in bipartite graphs. In target tracking scenarios, it mainly solves data association problems, i.e., associating and matching target boxes and detection boxes.

If we understand the bipartite graph as all detection boxes in two consecutive video frames, with the set of all detection boxes in the first frame called U and that in the second frame called V, it is easy to associate solving data association problems with the Hungarian algorithm for bipartite graphs. Since different detection boxes in the same frame will not correspond to the same target, there is no need for mutual association, while detection boxes in adjacent frames need to be connected. Finally, detection boxes in adjacent frames should be matched in pairs as well as possible. The optimal solution to this problem requires the Hungarian algorithm. The overall process is: traverse all detection boxes and tracking boxes and match them. If the match is successful, retain it; delete unmatched tracking boxes.

**5.4 Combination of Target Recognition and DeepSORT Algorithm**

As mentioned in previous chapters, the SORT algorithm mainly combines the Kalman filtering algorithm and the Hungarian algorithm. The DeepSORT algorithm, based on SORT, introduces a cascaded matching strategy and an enhanced state estimation mechanism. In actual scenarios, obstacles may occlude targets. When a target is occluded, the tracked target may fail to match the detected target in the current frame because the target temporarily leaves the field of view or is occluded by other objects. When these occluded targets reappear in the image, to reduce the number of target ID switches (ID-Switch), their IDs should remain unchanged. In other words, even if targets cannot be detected for a period, their previous IDs should continue to be used when they reappear to maintain tracking continuity and consistency. This avoids unnecessary changes in target IDs due to temporary occlusion, thereby improving the stability and accuracy of target tracking.

Combine target recognition with tracking algorithms to achieve tracking and gazing. The specific process is shown in Figure 13. [16] The output of the improved YOLOv5 algorithm introduced in previous chapters (feature maps with YOLO detection boxes) is used as input to the DeepSORT algorithm. First, the detection boxes enter the cascaded matching stage to determine whether the tracking boxes and detection boxes match. If the match is successful, the tracker updates based on the matching result; if not, it enters the IOU matching of the Hungarian algorithm: perform one-to-one IOU matching between the current frame's target detection boxes and the boxes predicted by Kalman filtering in the previous frame, then calculate the cost matrix based on the IOU matching results. The cost matrix is used as input to the Hungarian algorithm to obtain linear matching results. If the detection boxes and tracking boxes match in this stage, update the Kalman filter and then the tracker; unmatched tracking trajectories are considered lost, and the tracking boxes are deleted. The tracker chain uses Kalman filter prediction: if it is a confirmed state tracking box, it enters cascaded matching; if it is an unconfirmed tracking box, it enters IOU matching.

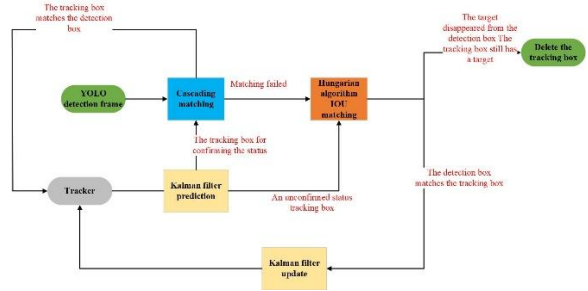

Figure 13 Flow Chart of Target Recognition and Tracking Algorithm

The DeepSORT algorithm optimizes target matching by applying the Hungarian algorithm to maintain tracking continuity. It also incorporates a motion model to predict target movement trends, estimating their future positions based on current positions and speeds. This prediction mechanism enhances tracking robustness, especially when targets temporarily leave the field of view or are occluded. DeepSORT can continuously track targets throughout the video sequence even when they are temporarily invisible. When targets reappear, DeepSORT uses their appearance features for matching, ensuring accurate recognition and re-association to maintain tracking continuity.

**6 Experimental Results and Analysis**

To verify the effectiveness of the proposed millisecond-response tracking and gazing system in reducing response delay and achieving stable tracking effects, target tracking and gazing experiments were conducted on the system.

**6.1 Introduction to Experimental Environment**

Experiments were conducted in an environment simulating actual road scenes, providing an open space for UAV operation. The UAV is equipped with advanced image sensors and computer vision technology, capable of real-time capturing vehicle images and analyzing their features. By identifying vehicle shapes, colors, and other features, the UAV can accurately track the target vehicle's path. This experimental environment can test the UAV's ability to accurately recognize and track specific vehicles in complex real environments.

The experiment uses video stream data captured in real-time by the Shanghai Bo UAV PTZ camera of a car model moving on a simulated road as input. The video resolution is 1920×1080, frame rate 30fps, with RGB three channels. The system uses Phytium's FT-2000/4 series processor and domestic Cambricon's MLU220 series intelligent accelerator card as the acceleration unit.

**6.2 Analysis of Experimental Results**

The camera continuously sends captured video data to the system's processing unit in the form of video frames through sensors. The system statistically analyzes the time consumption of each stage of single-frame image processing to obtain the overall processing delay.

The time for the system to process each frame of image is mainly divided into three parts: processing time on the CPU, processing time on the MLU220 accelerator card, and data transmission time between the CPU and MLU220. Processing on the CPU mainly involves adjusting and segmenting image size and shape, and being responsible for tracking and gazing functions after target recognition. The target recognition algorithm YOLOv5 is mainly accelerated for training and inference on the MLU220. Therefore, the experiment measured the processing time of 10 frames of images, comparing the processing time of each frame by the Cambricon MLU220 accelerator card before and after optimization. As can be seen from Figure 14, before algorithm optimization, the time for MLU220 to process one frame of image was almost above 41.5ms; after optimization, the processing time can be reduced to below 40.2ms. In addition, the data fluctuation before optimization was large, while after optimization, the fluctuation was small, indicating that the optimized system is more stable in processing video frames.

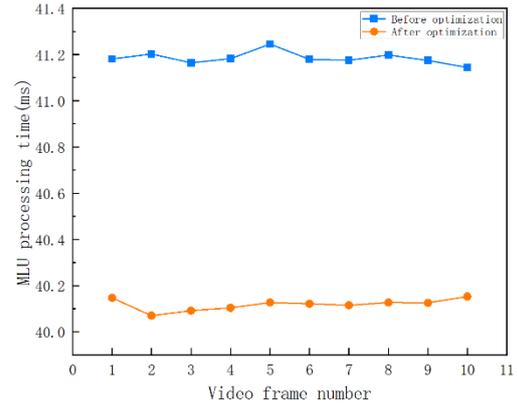

Figure 14 MLU processing time before and after optimization

The total time for processing image frames is determined by both CPU and MLU processing times. Optimization of CPU processing time mainly focuses on the target tracking part. Optimizing both parts significantly reduces the total processing time. As can be seen from Figure 15, the total time after optimization is significantly lower than before, indicating that the adopted optimization measures have significantly reduced the total system processing time. The data before optimization fluctuated greatly, ranging from 250ms to 450ms; after optimization, the average total system processing time decreased to 75ms (standard deviation ±15ms), a 65% reduction compared to before optimization. The fluctuation is small and relatively stable.

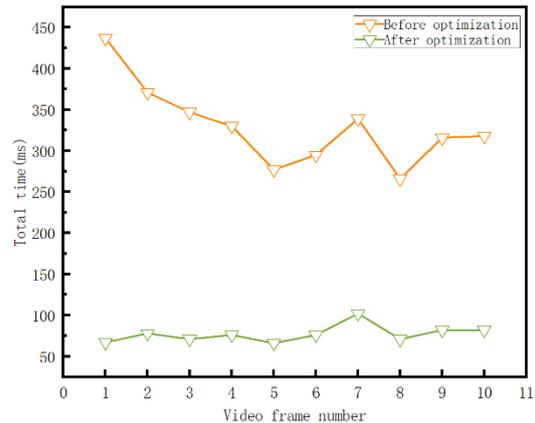

Figure 15 Total processing time before and after optimization

In the experiment, the recognition rates of 900, 1800, 2700, 3600, and 4500 image frames captured within 30s, 60s, 90s, 120s, and 150s were counted respectively. As shown in Figure 16, with the increase in the number of recognized image samples, the recognition accuracy shows a slow downward trend but remains at a high level, all above 98.5%. The reason why the recognition rate of twice the number of image frames is not equal to the square of the former is analyzed as follows: the target car travels along a specific circular route. It takes more than 90 seconds to complete one lap. Since the surrounding environment (background) of each section of the route within one lap is different, and some places have occlusions, the detection effect of video frames in two sections of the lap will differ, resulting in this phenomenon.

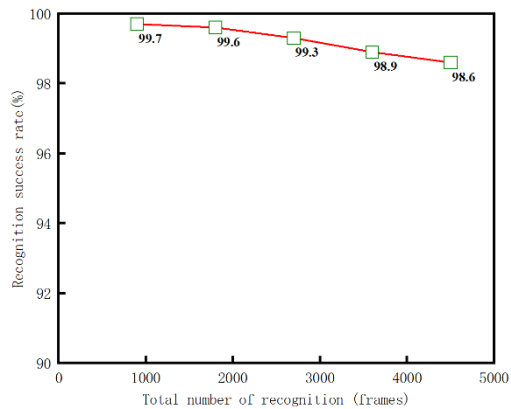

Figure 16 Change chart of recognition rate with total recognition count

In the experiment, the tracking effect of a single image frame at four scaling scales was tested within the recognizable target scale range, as shown in Figure 17. It can be seen from the figure that within a certain range, for the same image, as the scaling ratio increases, the recognition success rate significantly improves, meaning that the model can better capture the features of the recognized target at higher scaling ratios.

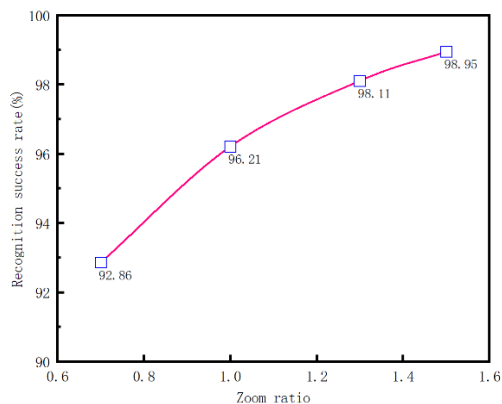

Figure 17 The graph of recognition rate changing with scaling ratio

# 7 Conclusion

This study proposes a millisecond-response UAV tracking and gazing system based on the domestic "Phytium + Cambricon" technical solution to address the weak target tracking capability and slow response speed of traditional PTZ cameras. The system's hardware includes domestic Phytium processors, Cambricon AI acceleration chip MLU220, and Shanghai Bo PTZ cameras responsible for perception. First, the PTZ camera transmits perceived video data to the processor, then the YOLOv5s model accelerated by MLU220 performs target detection and recognition, and further controls the PTZ camera to rotate to continuously recognize the target, achieving target recognition and tracking effects. The system improves computing power through multi-accelerator card parallel computing and reduces computational complexity through lightweight algorithm optimization, ultimately achieving stable tracking effects and millisecond-level response performance indicators, effectively solving response delay issues in target tracking and gazing. This study not only provides a solution for the development of UAV tracking and gazing technology but also offers empirical support for the application of domestic processors and AI acceleration chips in complex computing tasks. With technological progress and increasing environmental complexity (such as target occlusion by obstacles), there will be an urgent need for more intelligent and efficient computing power. The surge in data volume caused by complex environments may affect system response speed. Therefore, in future work, we will continue to optimize existing algorithms, aiming to enhance the system's real-time processing capability and intelligence level.